\documentclass[lettersize,journal]{IEEEtran}
\usepackage{amsmath,amsfonts}
\usepackage{algorithmic}
\usepackage{algorithm}
\usepackage{array}
\usepackage{makecell} 
\usepackage[caption=false,font=normalsize,labelfont=sf,textfont=sf]{subfig}
\usepackage{textcomp}
\usepackage{stfloats}
\usepackage{url}
\usepackage{verbatim}
\usepackage{booktabs}
\usepackage{multirow}
\usepackage{graphicx}
\usepackage{cite}
\usepackage{hyperref}   
\usepackage[table]{xcolor} 
\usepackage{pifont}         
\newcommand{\cmark}{\textcolor{green!60!black}{\ding{51}}}
\newcommand{\xmark}{\textcolor{red}{\ding{55}}}
\usepackage{amssymb}

\begin{document}

\title{PoreTrack3D: A Benchmark for Dynamic 3D Gaussian Splatting in Pore-Scale Facial Trajectory Tracking}

\author{
Dong Li,
Jiahao Xiong,
Yingda Huang,
and Le Chang*%
\thanks{*Corresponding author: lechang@gdut.edu.cn.}
\thanks{School of Automation, Guangdong University of Technology, Guangzhou, China. 
Emails: dong.li@gdut.edu.cn, 2112304352@mail2.gdut.edu.cn, 2112404370@mail2.gdut.edu.cn, lechang@gdut.edu.cn.}
}



\maketitle

\begin{abstract}
We introduce PoreTrack3D, the first benchmark for dynamic 3D Gaussian splatting in pore-scale, non-rigid 3D facial trajectory tracking. It contains over 440,000 facial trajectories in total, among which more than 52,000 are longer than 10 frames, including 68 manually reviewed trajectories that span the entire 150 frames. To the best of our knowledge, PoreTrack3D is the first benchmark dataset to capture both traditional facial landmarks and pore-scale keypoints trajectory, advancing the study of subtle facial expressions through the analysis of subtle skin-surface motion. We systematically evaluate state-of-the-art dynamic 3D Gaussian splatting methods on PoreTrack3D, establishing the first performance baseline in this domain. Overall, the pipeline developed for this benchmark dataset's creation establishes a new framework for high-fidelity facial motion capture and dynamic 3D reconstruction. Our dataset are publicly available at: \url{https://github.com/JHXion9/PoreTrack3D}
\end{abstract}

\section{Introduction}

Dynamic 3D reconstruction and trajectory tracking of non-rigid surfaces remain fundamental yet highly challenging tasks in computer vision and graphics \cite{mollahosseini2017affectnet, tellamekala20233d}. Applications such as photorealistic digital humans \cite{wang2021facial}, immersive human–computer interaction \cite{mollahosseini2017affectnet}, precision medical imaging \cite{kim2022numerical, davison2016samm}, and digital entertainment \cite{bao2021high} all demand accurate modeling of dense surface motion. Recent breakthroughs in static scene modeling, including Neural Radiance Fields (NeRF) \cite{mildenhall2021nerf} and 3D Gaussian Splatting \cite{kerbl20233d}, have demonstrated remarkable success in high-fidelity reconstruction. However, extending these advances to dynamic and highly deformable objects remains a significant open challenge.

A natural way to capture surface dynamics is through point-wise tracking of trajectories over time. In the 2D image domain, this paradigm has proven highly effective, with tasks like optical flow and the more recent long-term Tracking-Any-Point (TAP) leading to significant advancements. Pioneering benchmarks such as TAP-Vid \cite{doersch2022tap}, BADJA \cite{biggs2018creatures}, and JHMDB \cite{jhuang2013towards} have provided robust evaluation platforms, demonstrating the utility of 2D point tracking in applications ranging from video editing\cite{yu2023videodoodles} to robotics\cite{vecerik2024robotap}. Yet, 2D point tracking remains inherently limited to the image plane, failing to capture out-of-plane motion and thus providing only a partial measure of real-world dynamics.

\begin{figure}[ht!] 
    \centering
    \includegraphics[width=\linewidth]{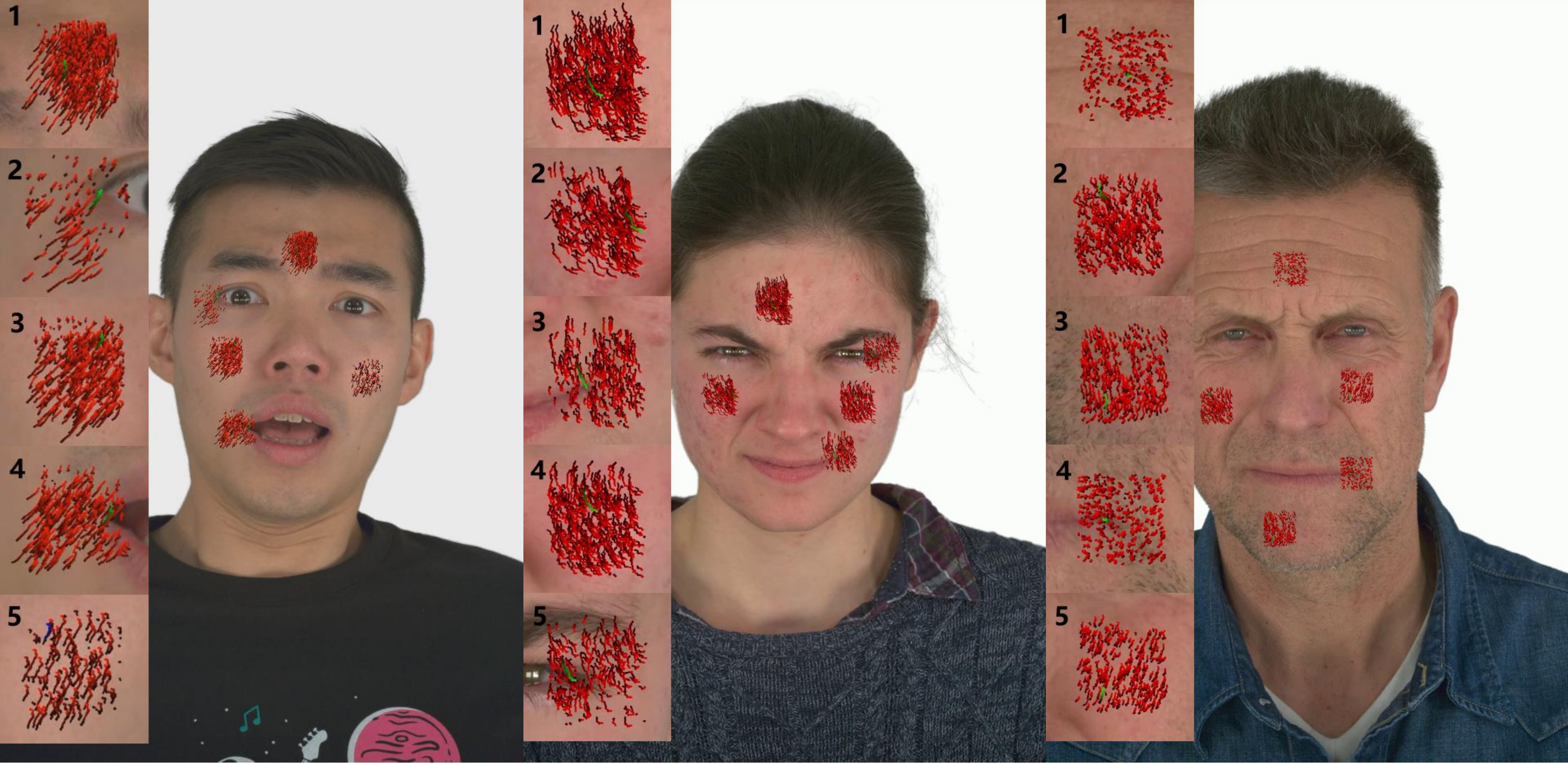}
    \caption{Representative Samples and Trajectory Visualizations of PoreTrack3D. For each subject, the locally magnified views are arranged in a counterclockwise order starting from the forehead region. To facilitate visual inspection, the trajectories of different types are color-coded: green denotes the primary trajectories that are successfully tracked across time, blue indicates the completion-derived trajectories obtained through temporal completion, and red represents supporting trajectories derived from neighboring keypoints. The direction of each trajectory is indicated by its color gradient, transitioning from black to the corresponding color (red, green, or blue). For clarity in visualization, only the trajectories over the previous 10 frames (out of the total 150) are shown for each displayed frame, and subsequent visualizations follow the same convention.
    }
    \label{Fig.1}
\end{figure}

Recognizing this limitation, recent efforts have extended tracking into 3D. Early attempts such as scene flow \cite{vedula1999three} generalized optical flow to three dimensions but were restricted to instantaneous motion without assessing long-term, occlusion-aware associations. More recent datasets, including TAP-3D \cite{koppula2024tapvid} and Stereo4D \cite{jin2024stereo4d}, have begun to define evaluation standards for general 3D motion. However, these benchmarks primarily focus on coarse, keypoint-level trajectories of rigid or semi-rigid objects, and thus fall short of capturing the subtle, textureless deformations observed in highly non-rigid surfaces. Their spatial resolution and motion precision—typically at the centimeter level—are insufficient to evaluate methods like Dynamic 3D Gaussian Splatting methods, which aim to model dense, continuous, and sub-millimeter geometric motion. Recent studies \cite{luiten2024dynamic, duisterhof2023deformgs, zhang2024dynamic} have further demonstrated that Dynamic 3D Gaussian Splatting methods achieve superior temporal coherence and tracking accuracy compared with correspondence-based tracking methods—such as PIPS++~\cite{zheng2023pointodyssey}, CoTracker~\cite{karaev2024cotracker}, and SpatialTracker~\cite{xiao2024spatialtracker}—by explicitly modeling scene geometry and motion within a unified 3D representation. This geometric consistency allows them to better capture continuous deformations and subtle surface motions, underscoring the need for a dedicated benchmark tailored to such subtle, non-rigid dynamics.

Among such non-rigid objects, the human face presents a particularly challenging case. Facial dynamics are driven by complex muscle activations that induce subtle yet critical deformations—such as the protrusion of lips, the strain of skin during a smile, or the emergence of fine wrinkles. Parametric 3D morphable models (3DMMs) \cite{blanz2023morphable}, such as FLAME \cite{li2017learning}, have provided a compact representation of facial geometry and expressions, and are widely used to synthesize motion data in controlled settings. However, datasets generated from such parametric priors—like FacialFlowNet \cite{lu2024facialflownet}—rely on synthetic renderings with non-realistic textures and do not provide explicit 3D trajectories. As a result, they fail to reflect the true physical behavior of facial surfaces at fine spatial scales, leaving the dense, pore-scale motion of real human faces largely unexplored.

To fill this gap, we introduce PoreTrack3D, the first pore-scale 3D facial point trajectory benchmark dataset designed for high-precision facial dynamics analysis. Representative samples and trajectory visualizations are shown in Fig. \ref{Fig.1}. Our main contributions are summarized as follows:

\begin{itemize}
    \item PoreTrack3D is introduced as the first benchmark for evaluating dynamic 3D Gaussian splatting methods in pore-scale, non-rigid facial trajectory tracking. In contrast to existing 2D or coarse 3D tracking benchmarks, PoreTrack3D captures both facial landmark trajectories and pore-scale trajectories, enabling quantitative assessment of subtle, non-rigid facial dynamics.
    
    \item A comprehensive data creation framework is developed, integrating automated trajectory extraction with interactive manual review. This framework supports the generation of high-precision 3D trajectories while maintaining both accuracy and scalability.
    
    \item We establish a new evaluation benchmark for dynamic 3D facial motion. Leveraging PoreTrack3D, we present the first subtle motion-level assessment of state-of-the-art dynamic 3D Gaussian splatting methods, revealing their strengths and weaknesses in capturing high-fidelity surface dynamics.
\end{itemize}

\section{Related  Work}\label{sec:related}

\subsection{Motion Data}

\begin{table*}[h!]
\centering
\caption{Overview of 3D Point Trajectory Tracking datasets}
\label{tab:table1}

\renewcommand{\arraystretch}{2}

\resizebox{\textwidth}{!}{%
\begin{tabular}{lccccccccc}
\toprule
\textbf{Dataset} & 
\makecell{\textbf{Real-World} \\ \textbf{Data}} &
\makecell{\textbf{Trajectory} \\ \textbf{Manual Review}} &
\makecell{\textbf{High-Res.} \\ \textbf{Face}} &
\textbf{Mesh} &
\textbf{Textureless} &
\makecell{\textbf{Testing Set} \\ \textbf{\#Trajectories}} &
\makecell{\textbf{Testing Set} \\ \textbf{Avg. Length (Frames)}} &
\textbf{\#Views} &
\textbf{$\delta$-Threshold in Evaluation} \\
\midrule

TAP-3D-ADT \cite{koppula2024tapvid}   
& \xmark & \xmark & \xmark & \cmark & \cmark & 1951K & 300 & 1-2 
& \makecell{$\delta_{2D}=\{1,2,4,8,16\}$ pixel \\ unprojected to depth-relative 3D error} \\

TAP-3D-DriveTrack \cite{koppula2024tapvid}   
& \cmark & \xmark & \xmark & \xmark & \cmark & 616K & 25-300 & 5 
& \makecell{$\delta_{2D}=\{1,2,4,8,16\}$ pixel \\ unprojected to depth-relative 3D error} \\

TAP-3D-Panoptic Studio \cite{koppula2024tapvid}   
& \cmark & \xmark & \xmark & \xmark & \cmark & 5.3K & 150 & 480 
& \makecell{$\delta_{2D}=\{1,2,4,8,16\}$ pixel \\ unprojected to depth-relative 3D error} \\

Stereo4D \cite{jin2024stereo4d}  
& \cmark & \xmark & \xmark & \xmark & \xmark & - & - & 2 
& $\delta_{3D}=\{5, 10\}$ cm \\

Argoverse 2 \cite{wilson2023argoverse}        
& \cmark & \xmark & \xmark & \xmark & \xmark & 1390K & 110 & 9 
& - \\

3DPW \cite{von2018recovering}   
& \cmark & \xmark & \xmark & \cmark & \xmark & - & - & 1 
& - \\

CMU Panoptic Studio \cite{joo2015panoptic} 
& \cmark & \xmark & \xmark & \xmark & \cmark & - & - & 480 
& - \\

Dynamic FAUST \cite{bogo2017dynamic} 
& \cmark & \xmark & \xmark & \cmark & \cmark & 129 & 300 & 44 
& - \\

PanopticSports \cite{luiten2024dynamic} 
& \cmark & \cmark & \xmark & \xmark & \xmark & 21 & 150 & 27 
& $\delta_{3D}=\{1, 2, 4, 8, 16\}$ cm \\

\textbf{PoreTrack3D} 
& \cmark & \cmark & \cmark & \cmark & \cmark & 68 & 150 & 16 
& $\delta_{3D}=\{1, 1.5, 2, 2.5\}$ mm \\
\bottomrule
\end{tabular}%
}
\end{table*}

Large-scale, high-fidelity motion datasets are essential for advancing and evaluating 3D dynamic vision algorithms. In general dynamic scenes, a wide range of benchmarks has been established. For example, MPI-Sintel \cite{butler2012naturalistic} provided dense optical flow fields, advancing 2D non-rigid motion estimation. More recent 2D synthetic resources such as TAP-Vid-Kubric \cite{doersch2022tap} and PointOdyssey \cite{zheng2023pointodyssey} leveraged physics engines to generate large collections of dense point trajectories. Building on the TAP paradigm, TAP-3D \cite{koppula2024tapvid} extended point tracking evaluation into 3D by establishing benchmarks that measure the trajectories of arbitrary points in 3D space. It includes both synthetic subsets (e.g., TAP-3D-ADT, derived from simulated environments) and real-world subsets (e.g., DriveTrack, Panoptic Studio), offering broad coverage across domains. In parallel, Stereo4D \cite{jin2024stereo4d} learned 3D motion directly from Internet stereo videos, providing large-scale, in-the-wild pseudo-ground-truth trajectories. Large-scale driving datasets such as Argoverse 2 \cite{wilson2023argoverse} further captured complex multi-agent interactions in urban environments.

Within the broader landscape of motion datasets, human motion capture has emerged as a particularly active research area, with most efforts centered on full-body dynamics. Human3.6M \cite{ionescu2013human3}, a foundational dataset, employed high-precision MoCap systems to record 3D skeletal joint trajectories in controlled environments, setting a standard for pose estimation. The CMU Panoptic Studio \cite{joo2015panoptic} scaled this paradigm with over 500 synchronized cameras, enabling detailed multi-person 3D reconstruction. Subsequent datasets extended the scope to dense, non-rigid tracking: Dynamic FAUST \cite{bogo2017dynamic} captured dynamic body shapes with depth sensors, and PanopticSports \cite{luiten2024dynamic} introduced athletic sequences with challenging surface deformations. Meanwhile, 3DPW \cite{von2018recovering} pushed research into in-the-wild scenarios by combining wearable IMUs and cameras to capture natural 3D poses.

In spite of recent progress, existing datasets share a key limitation: they primarily capture large-scale motion. While effective for centimeter-level movements of bodies and objects, this resolution is insufficient for the human face. Facial expressions arise from millimeter skin deformations driven by underlying muscles. The absence of datasets that capture such subtle dynamics remains a critical bottleneck for facial analysis research, hindering progress and the development of more precise algorithms. We overcome this limitation by introducing PoreTrack3D, a high-resolution facial dataset with millimeter-level 3D pore-scale trajectories that support precise analysis of fine facial dynamics. Table~\ref{tab:table1} reports only datasets that provide 3D trajectories, as this requirement is fundamental for a direct and fair comparison; works lacking 3D supervision are therefore omitted.

\subsection{Dynamic 3D Gaussian Splatting Methods for Tracking and Reconstruction}

Dynamic 3D Gaussian Splatting methods have recently emerged as a powerful framework for joint scene reconstruction and motion tracking\cite{luiten2024dynamic, duisterhof2023deformgs, zhang2024dynamic, kerbl20233d}. Compared with traditional NeRF\cite{mildenhall2021nerf} or mesh-based dynamic reconstruction methods\cite{grassal2022neural, li2009robust}, Dynamic 3D Gaussian Splatting methods achieve superior rendering fidelity and enables dense, temporally consistent tracking of surface points in real time. By explicitly modeling scene geometry as collections of anisotropic Gaussian primitives, these methods inherently support continuous motion representation and millimeter trajectory estimation—capabilities that are challenging for conventional volumetric or parametric approaches. Recent evaluations \cite{luiten2024dynamic, duisterhof2023deformgs, zhang2024dynamic} further indicate that Dynamic 3D Gaussian Splatting methods surpass correspondence-based tracking methods in temporal coherence and motion accuracy, reinforcing their suitability as the primary focus of our benchmark.

Recently works such as Dynamic 3D Gaussians (Dyn3DGS)\cite{luiten2024dynamic} pioneered this paradigm by frame-wise optimizing Gaussian attributes, achieving high-quality rendering and explicit per-point tracking. Dyn3DGS-robot\cite{zhang2024dynamic} extended this concept by introducing graph neural networks over sparse control particles to model predictive dynamics. 4D Gaussian Splatting (4DGS)\cite{wu20244d} further improved efficiency through a deformation-field formulation that maps a canonical set of Gaussians across time, significantly reducing storage and computation while preserving temporal correspondence.

Subsequent methods continued to enhance the temporal modeling of Gaussians. DeformGS\cite{duisterhof2023deformgs}, Gaussian-Flow\cite{lin2024gaussian}, and DynMF\cite{kratimenos2024dynmf} explicitly learned deformation or motion bases to ensure temporally coherent Gaussian trajectories, thereby supporting both tracking and reconstruction. In contrast, Compact-D3D\cite{katsumata2024compact} focused primarily on compact dynamic scene representation: it models temporal variations using Fourier components without maintaining consistent Gaussian identities across frames. Consequently, while Compact-D3D achieves efficient dynamic reconstruction, it cannot produce explicit per-point trajectories for motion tracking.

Despite their effectiveness on general dynamic scenes, these approaches face new challenges when applied to highly non-rigid objects such as human faces. In such cases, the lack of face-specific priors leads to degraded tracking accuracy and robustness under weak texture or rapid expression changes.

\subsection{Face-Specific Dynamic 3D Gaussian Splatting Methods for Reconstruction and Tracking}

For human faces, researchers have introduced parametric models as strong priors, achieving notable progress. Parametric models such as FLAME \cite{li2017learning} provided a low-dimensional control space for identity, expression, and pose, constraining the reconstruction space to better align with facial anatomy.  

GaussianAvatars (GSAvatars) \cite{qian2024gaussianavatars} exemplified this direction by binding 3D Gaussians to FLAME mesh triangles. The fixed topology of FLAME ensures Gaussian motion is consistent with facial deformations, enabling high-fidelity and controllable facial animation. This approach elegantly combines the controllability of parametric models with the rendering expressiveness of 3D Gaussians. Because each Gaussian inherits its motion from the underlying mesh vertices, their temporal evolution naturally forms coherent trajectories across frames. This makes GSAvatars particularly suitable for tracking and analyzing consistent 3D motion at the vertex or region level, without suffering from correspondence ambiguity. Its structured representation preserves temporal coherence while maintaining fine geometric detail, offering a strong balance between model interpretability and visual fidelity.

FastAvatar \cite{wu2025fastavatar}, GPAvatar \cite{feng2025gpavatar}, and Gaussian Déjà-vu \cite{Yan_2025_WACV} further improve the efficiency and fidelity of 3D Gaussian-based facial reconstruction. These methods replace per-frame Gaussian optimization with a neural canonical-space representation, where a network—typically an MLP, Transformer, or GNN—predicts a shared set of Gaussians from multi-view or multi-frame observations. By aggregating information into a unified canonical field, they achieve fast and high-quality avatar generation. However, since the Gaussian identities are not preserved across frames, these canonical-space or graph-based representations lack explicit temporal correspondence and thus cannot provide point-level tracking or subtle motion analysis.

Despite these advances, existing face-specific 3D Gaussian Splatting methods mainly focus on improving reconstruction fidelity and visual realism, while neglecting temporal geometric consistency. None of these approaches explicitly establish or evaluate point-level correspondences across time, making it difficult to quantify subtle facial dynamics such as micro-expressions, skin stretching, or pore-scale motion. Moreover, the absence of a standardized benchmark prevents fair and reproducible evaluation of temporal accuracy at millimeter precision. To address this gap, we propose PoreTrack3D, a benchmark specifically designed for high-precision facial dynamic analysis, offering millimeter-level 3D trajectories of high-resolution faces for evaluating subtle motion and tracking consistency in dynamic 3D Gaussian representations.

\section{PoreTrack3D}

\begin{figure*}[ht!] 
    \centering
    \includegraphics[width=\linewidth]{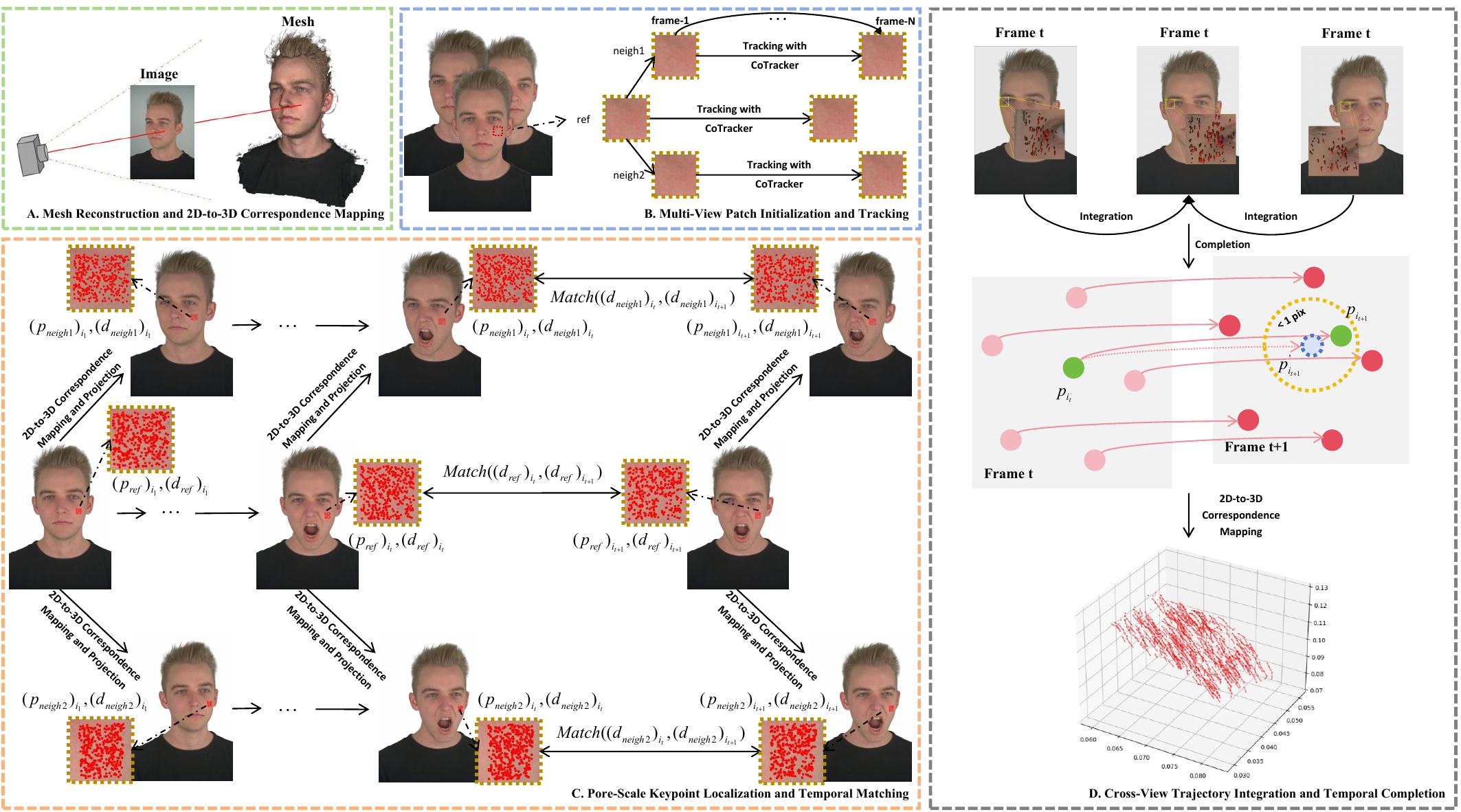}
    \caption{Overview of the proposed four-stage multi-view trajectory reconstruction framework: (A) Mesh reconstruction and 2D-to-3D correspondence mapping, (B) Multi-view patch initialization and tracking, (C) Pore-scale keypoint localization and temporal matching, and (D) Cross-view trajectory integration and temporal completion.
    }
    \label{Fig.2}
\end{figure*}

For our benchmark, which requires extremely high-fidelity facial geometry, sufficient multi-view coverage, and diverse non-rigid expressions, we build upon the NeRSemble dataset \cite{kirschstein2023nersemble} as a high-quality data source. PoreTrack3D contains 128 multi-view facial sequences from 8 subjects, each recorded with a synchronized 16-camera high-resolution (7.1-megapixel) setup, enabling detailed capture of facial dynamics across diverse expressions.

Leveraging its rich capture setting, a four-stage multi-view trajectory reconstruction framework (Fig. \ref{Fig.2}) is designed to reconstruct and track subtle, non-rigid surface trajectories across frames. The framework consists of four stages: (A) Mesh reconstruction and 2D-to-3D correspondence mapping, (B) Multi-view patch initialization and tracking, (C) Pore-scale keypoint localization and temporal matching, and (D) Cross-view trajectory integration and temporal completion. The technical details of each stage are elaborated below. Note that the 2D-to-3D correspondence mapping is defined in stage (A) and subsequently applied throughout stages (B)–(D).

\subsection{Mesh Reconstruction and 2D-to-3D correspondence Mapping}
To obtain accurate geometric priors for subsequent 2D-to-3D correspondence mapping, we reconstruct a high-fidelity facial mesh by refining the standard COLMAP\cite{schoenberger2016sfm, schoenberger2016mvs} pipeline. Specifically, we employ the PSIFT algorithm\cite{li2015design} to detect dense and stable keypoints, extract discriminative feature descriptors using Pore-HyNet-V4\cite{li2025porescale}, and apply GMS matching\cite{bian2017gms} for robust correspondence filtering. The resulting sparse correspondences are then used to estimate camera poses and generate dense point clouds, from which a high-quality facial mesh is obtained through Poisson reconstruction. 

Based on the facial mesh, we establish accurate correspondences between 2D keypoints and the 3D locations on the facial surface by casting rays from the camera center into the mesh surface. Let $R \in \mathbb{R}^{3\times3}$ and $T \in \mathbb{R}^{3\times1}$ denote the camera rotation and translation from the world coordinate system to the camera coordinate frame. The camera center in the world coordinate system is then given by
\begin{equation}
C_\text{world} = -R^{\top} T.
\end{equation}

\textbf{Ray construction from 2D keypoints.}  
For each 2D keypoint $p_i = (u_i, v_i)$, the corresponding direction in the camera coordinate frame under the pinhole model is
\begin{equation}
(d_\text{cam})_i =
\begin{bmatrix}
\frac{u_i - c_x}{f_x} \\
\frac{v_i - c_y}{f_y} \\
1
\end{bmatrix}.
\end{equation}
This direction is transformed to the world coordinate system as
\begin{equation}
(d_\text{world})_i =
\frac{R^{\top} (d_\text{cam})_i}{\| R^{\top} (d_\text{cam})_i \|}.
\end{equation}
The resulting ray is
\begin{equation}
r_i(s) = C_\text{world} + s \,(d_\text{world})_i,\quad s>0.
\label{eq:ray_world}
\end{equation}

\textbf{Ray–mesh intersection.}  
Intersections between $r_i(s)$ and the facial mesh are computed using an optimized ray–triangle intersection algorithm~\cite{wald2014embree}. The intersection parameter $s_i^*$ yields the 3D point
\begin{equation}
(P_\text{world})_i = C_\text{world} + s_i^* \,(d_\text{world})_i.
\label{eq:world_hit}
\end{equation}
The 3D position $(P_\text{world})_i$ is therefore associated with the 2D keypoint $(u_i, v_i)$, establishing the corresponding 2D-to-3D mapping for subsequent trajectory reconstruction.

\subsection{Multi-View Patch Initialization and Tracking}

After obtaining the 2D-to-3D correspondence mapping described in the previous section, we manually select facial landmarks and pore-scale keypoints from the reference (frontal) view, ensuring that the chosen regions provide diverse coverage of the facial surface. The centers of these regions, are first mapped to their corresponding 3D coordinates using the 2D-to-3D correspondence mapping method. Each 3D point is then projected onto its two neighboring camera views through their respective intrinsic and extrinsic parameters, forming multi-view correspondences for the same facial location.

These projected centers serve as initialization points for patch tracking. In each view, we crop a $100 \times 100$ pixel patch around the projected center and employ the CoTracker \cite{karaev2024cotracker} to track its motion across consecutive frames. This process provides coarse but stable motion tracking for each view, defining a temporally consistent local tracking window where pore-scale keypoints localization and feature matching are subsequently performed.

\subsection{Pore-Scale Keypoint Localization and Temporal Matching}

Within each tracked patch obtained from the previous step, we perform pore-scale analysis to extract dense and temporally consistent correspondences. This process involves three components:

\textbf{Keypoint Detection.}
We apply the PSIFT algorithm to each patch to detect precise pore positions, yielding a set of detected keypoints denoted as $p_{i_t}$.

\textbf{Descriptor Extraction.}
For each keypoint $p_{i_t}$, a $64 \times 64$ local patch is cropped, resized to $32 \times 32$, and processed by the Pore-HyNet-V4 network to generate discriminative feature descriptors $d_{i_t}$.

\textbf{Temporal Matching.}
For consecutive frames $t$ and $t+1$, a temporal correspondence matrix $M_{t}$ is constructed via descriptor-based matching:
\begin{equation}
    M_{t} = \text{Match}(d_{t}, d_{t+1}),
\end{equation}
where $\text{Match}(\cdot)$ applies nearest-neighbor search in descriptor space, followed by a ratio test (nearest/second-nearest $<0.85$) and RANSAC-based geometric verification to remove outliers.  In addition to automatic descriptor matching, $M_{t}$ also supports manually annotated correspondences, allowing human-supervised specification of reliable point-to-point matches when necessary.

The above three components are first applied to the reference (frontal) view, yielding reliable keypoints $(p_{\text{ref}})_{i_t}$ and their descriptors $(d_{\text{ref}})_{i_t}$ across all frames. These keypoints $(p_{\text{ref}})_{i_t}$ are then mapped to 3D via the 2D-to-3D correspondence mapping method and reprojected onto the two neighboring views using their respective camera parameters, producing ${(p_{\text{neigh1}})_{i_t}}$ and ${(p_{\text{neigh2}})_{i_t}}$.

Unlike the reference view, the neighboring views do not require keypoint detection. Instead, for each reprojected keypoint, a local patch is extracted to compute descriptors ${(d_{\text{neigh1}})_{i_t}}$ and ${(d_{\text{neigh2}})_{i_t}}$ using Pore-HyNet-V4. The same temporal matching procedure is then applied to these descriptors to enforce geometric and temporal consistency across views.

Finally, the resulting temporal correspondence matrices $(M_{view})_t$, $view \in {\text{ref}, \text{neigh1}, \text{neigh2}}$, record the valid keypoint associations between consecutive frames, forming the foundation for constructing a unified correspondence space that aligns temporally consistent keypoints across views. This unified space serves as the basis for subsequent trajectory integration and temporal completion.

\subsection{Cross-View Trajectory Integration and Temporal Completion}
\begin{figure}[t!] 
    \centering

    \includegraphics[width=\linewidth]{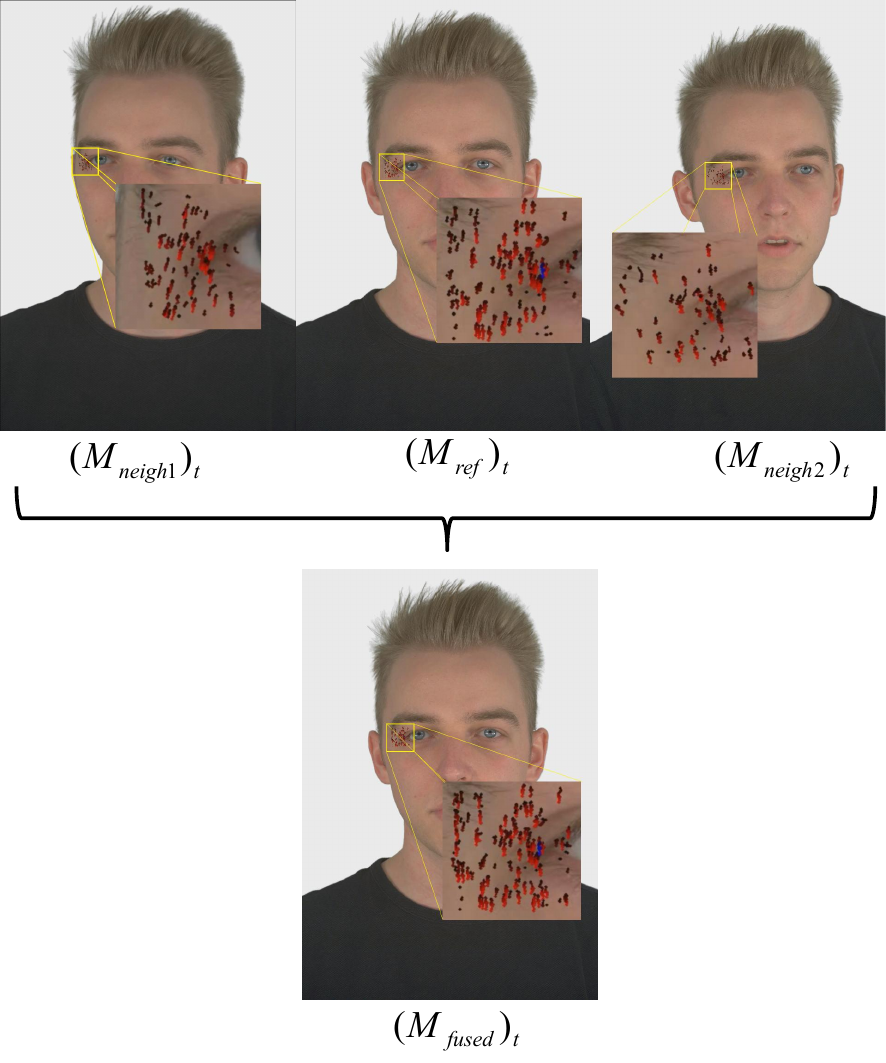}

    \caption{Cross-view trajectory integration, where missing matches in the reference view are supplemented using neighboring projections.} 

    \label{Fig.3}
\end{figure}

Based on the temporal and cross-view correspondences obtained in the previous stage, we integrate the matched keypoints across views at each time step to obtain spatially consistent multi-view observations. When a keypoint match is missing in the reference view, we retrieve the corresponding projected keypoint from other views and supplement the missing record using its matching results. The overall integration procedure is illustrated in Fig. \ref{Fig.3}, and can be formulated as

\begin{equation}
(M_{\text{fused}})_t(i)=
\begin{cases}
(M_{\text{ref}})_t(i), & \text{if } (M_{\text{ref}})_t(i) \text{ exists},\\[2pt]
(M_{\text{neigh1}})_t(i), & \text{else if } (M_{\text{neigh1}})_t(i) \text{ exists},\\[2pt]
(M_{\text{neigh2}})_t(i), & \text{else if } (M_{\text{neigh2}})_t(i) \text{ exists},\\[2pt]
\varnothing, & \text{otherwise,}
\end{cases}
\end{equation}

\noindent
where \( i \) denotes the index of a keypoint in the unified correspondence space, and \( \varnothing \) represents a missing temporal correspondence where the keypoint is not observed in frame $t$.

After cross-view integration, the trajectories may still contain discontinuities or missing points caused by inconsistent matching across frames. To recover these incomplete motion paths and maintain temporal coherence, we further apply a temporal completion process that refines and extends the integrated trajectories.

Specifically, we start by completing the primary trajectories. For pore-scale keypoints, the primary trajectories correspond to the longest (usually $>$70 frames) and most reliable motion sequences. For facial landmarks, the primary trajectories are defined by the landmark points detected on the first frame using PIPNet\cite{jin2021pixel}, together with their associated trajectories across the sequence. Suppose a point $p_{i_t}$ in a primary trajectory loses its correspondence between frame $t$ and $t{+}1$. To estimate its motion, we first select its five nearest neighboring keypoints that successfully maintain correspondences between frame $t$ and $t{+}1$, forming a locally reliable motion reference. The tentative position $p_{i_{t+1}}'$ is then predicted by computing a weighted average of their motion vectors, where the weight of each neighbor is inversely proportional to its Euclidean distance from the target point. We subsequently search within a 1-pixel radius of the predicted position $p_{i_{t+1}}'$ for any existing keypoint at frame $t{+}1$. If such a keypoint $p_{i_{t+1}}$ is found, it replaces the predicted position and the two trajectories are merged and treated as a single continuous sequence; otherwise, the predicted position $p_{i_{t+1}}'$ is used to complete the missing segment of the original trajectory, which we refer to as completion-derived trajectories.

After all primary trajectories are processed in the forward direction, the same completion procedure is performed backward over the entire sequence. This forward–then–backward two-pass completion extends each trajectory from its valid temporal span toward both ends of the sequence, yielding temporally consistent and continuous keypoint trajectories. Finally, the completed 2D trajectories are mapped to 3D using the 2D-to-3D correspondence mapping method, yielding the spatiotemporal motion trajectories of tracked keypoints, denoted as $\{P_i\}$.

\subsection{Manual Review}
\begin{figure}[t!] 
    \centering

    \includegraphics[width=\linewidth]{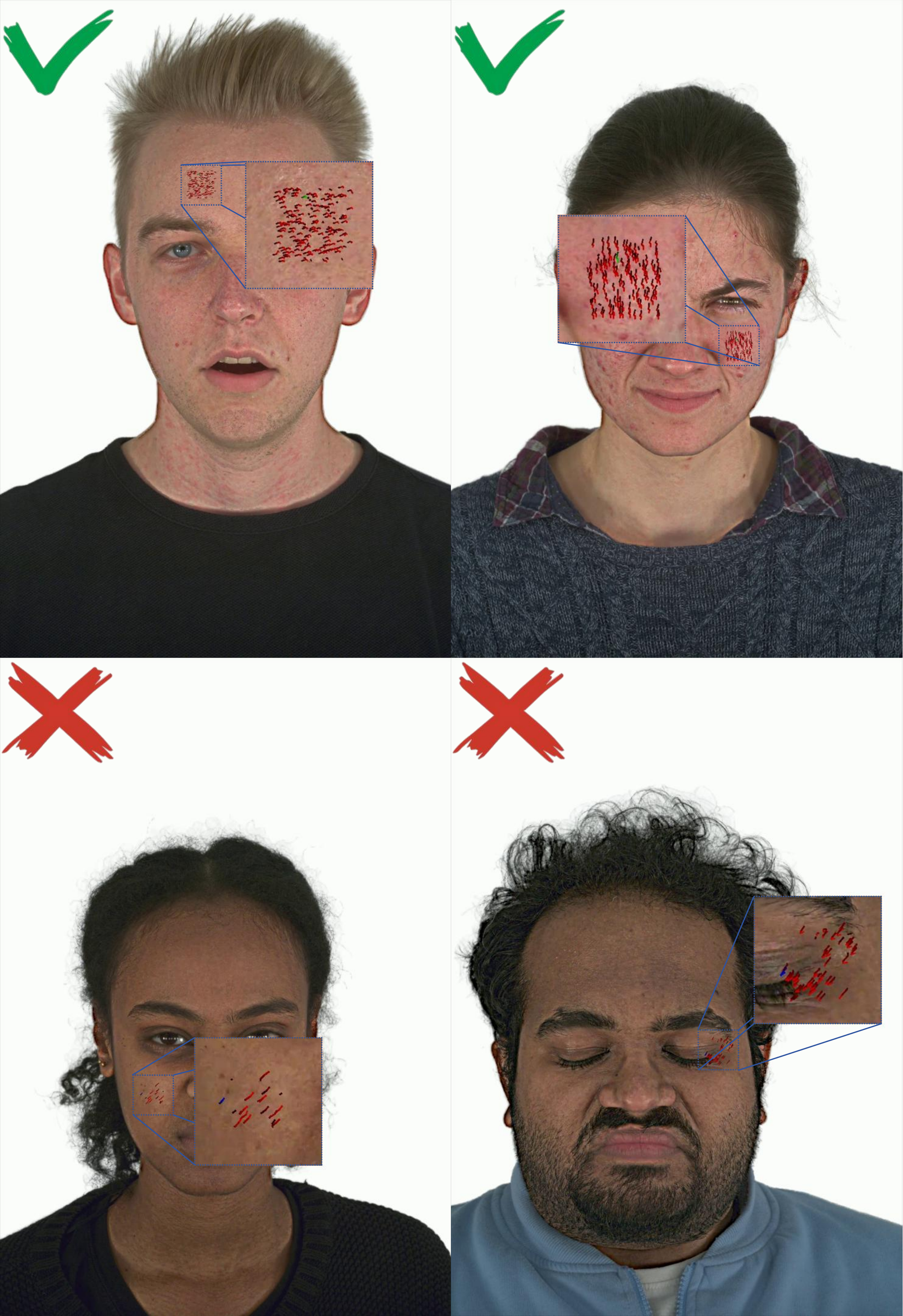}

    \caption{Manual review visualization of reconstructed trajectories on the texture-enhanced reference view. To illustrate representative failure cases, two subjects that failed manual review are shown here but are excluded from the final dataset. The synthetic texture enhancements do not reflect the subjects’ actual appearance and serve only to aid manual review.} 

    \label{Fig.4}
\end{figure}


To ensure the overall reliability and physical realism of the dataset, all reconstructed trajectories are manually reviewed for temporal consistency, geometric plausibility, and alignment with local texture motion in the reference view. A trajectory is regarded as valid when its motion trend follows the local texture motion and remains temporally coherent across frames. Small deviations within a 2 mm tolerance are considered acceptable and do not affect its validity, whereas trajectories exhibiting inconsistent motion patterns, spatial discontinuities, or visible drifts beyond this threshold are marked as invalid.

As shown in Fig.~\ref{Fig.4}, the reconstructed keypoint trajectories are projected onto the texture-enhanced reference view to facilitate intuitive inspection of motion trends and spatial coherence. The enhancement is achieved through a GMM-based texture refinement pipeline \cite{pun2014skin}, which detects facial landmarks, segments skin regions, and selectively amplifies subtle surface details to improve the visibility and contrast of motion traces. Trajectory colors encode their semantic roles: green denotes valid segments of the primary trajectories, blue indicates completion-derived trajectories used to fill temporal gaps, and red represents supporting trajectories derived from neighboring keypoints. A color gradient from black to the representative hue (green, blue, or red) indicates temporal direction, with darker colors corresponding to earlier frames. During manual review, only the green hollow circles are displayed to reduce visual clutter; the red supporting trajectories are included in the paper figures solely for illustration, providing spatial context and aiding in identifying potential inconsistencies.

For example, the first subject in the second row of Fig.~\ref{Fig.4} fails the review because the supporting red trajectories show inconsistent lengths, implying unstable local tracking quality, and the blue trajectory is significantly longer than its red neighbors, indicating a drift. The second subject in the second row fails due to eye closure, where the eyelid occludes the eye corner, causing the tracked point to drift upward toward the eyelid region in later frames.

\subsection{PoreTrack3D Dataset}
\begin{figure}[t!] 
    \centering
    \includegraphics[width=1\linewidth]{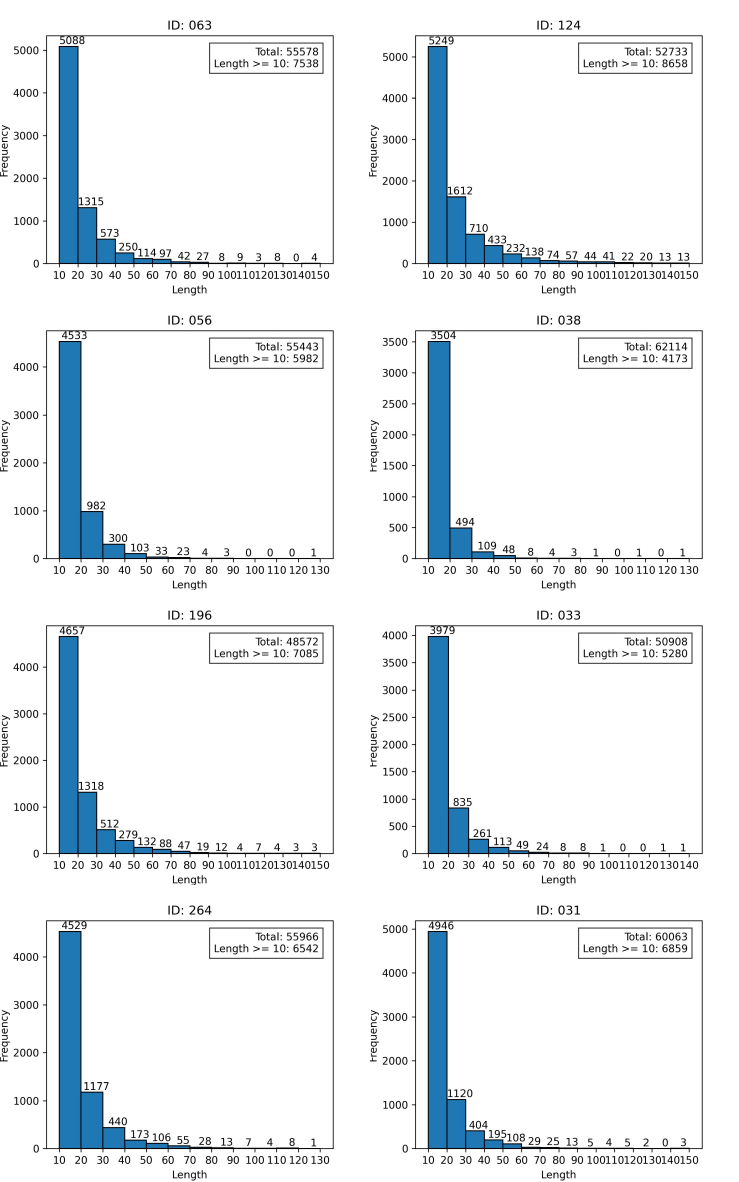}
    \caption{Histogram of trajectory statistics in the dataset, showing the total number of trajectories and those longer than 10 frames for each subject. Overall, trajectories exceeding 10 frames account for less than 15\% of the total.} 
    \label{Fig.5}
\end{figure}
\begin{figure*}[t!] 
    \centering
    \includegraphics[width=1\linewidth]{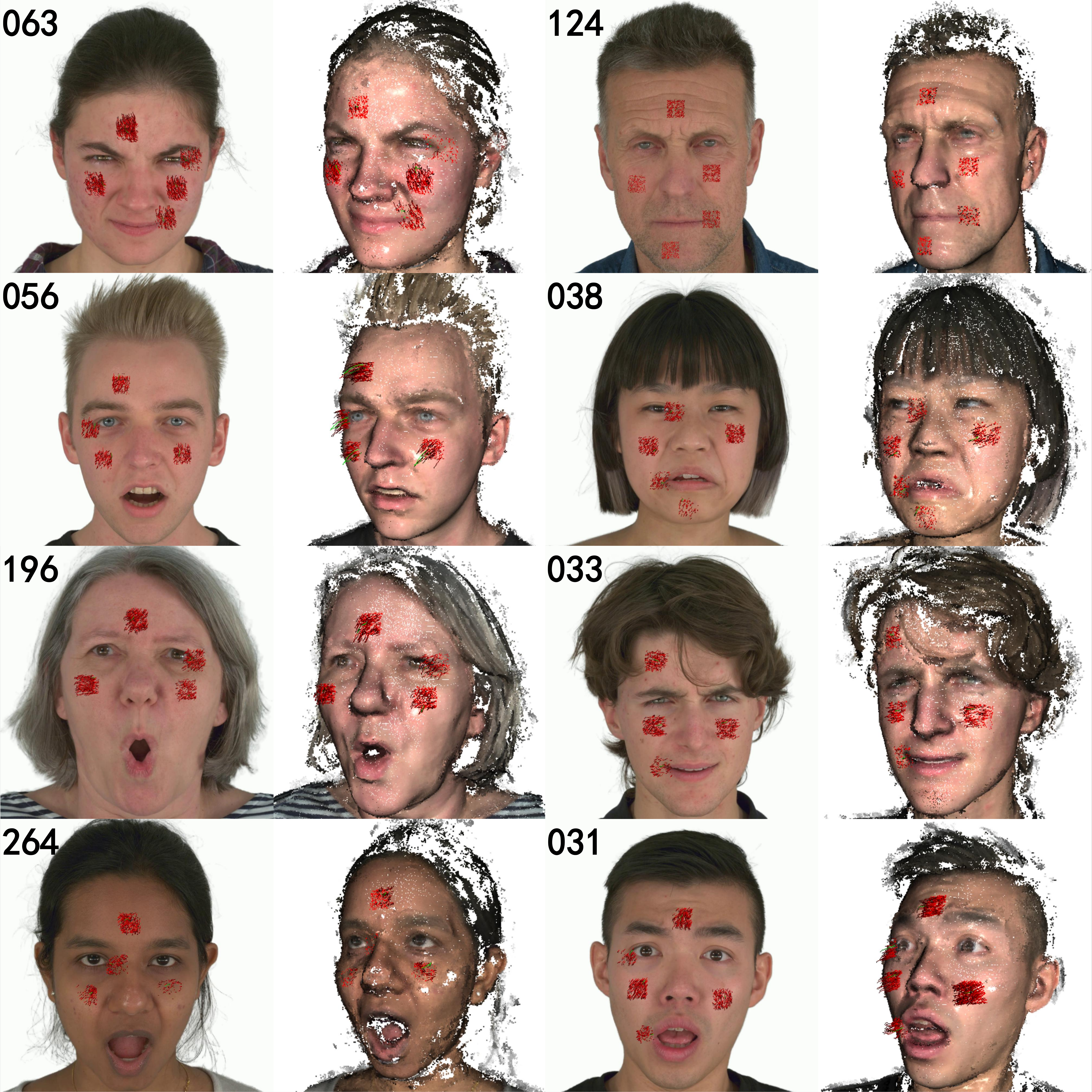}
    \caption{Visualization of 10-frame facial trajectories from PoreTrack3D, including both facial landmarks and pore-scale keypoints. For clarity, only 36 out of the 68 available trajectories are shown. The subjects are arranged in ascending order of task difficulty.} 
    \label{Fig.6}
\end{figure*}

The PoreTrack3D benchmark is built upon the publicly released NeRSemble dataset, which provides synchronized
multi-view facial sequences captured from 16 fixed viewpoints. For each subject, we extract a 150-frame subsequence
at a spatial resolution of $2200 \times 3208$, using the provided calibrated intrinsics and extrinsics for all views.

Based on these calibrated data, we reconstruct a 3D point cloud for the initial frame and generate dense temporal trajectories through our proposed four-stage trajectory reconstruction framework. In total, PoreTrack3D contains over 440,000 trajectories, among which more than 52,000 extend beyond 10 frames. Of these, 68 primary trajectories span the full 150 frames and have been manually reviewed, as illustrated in Fig.~\ref{Fig.5}. These trajectories cover both conventional facial landmarks and pore-scale surface keypoints, enabling quantitative analysis of subtle facial deformation. Specifically, among the 68 full-length trajectories, 20 trajectories correspond to facial landmarks and 48 trajectories correspond to pore-scale keypoints. Notably, trajectories longer than 10 frames constitute less than 15\% of the total, underscoring the challenge of achieving temporally coherent keypoint tracking under realistic motion and illumination variations.

Fig.~\ref{Fig.6} visualizes 10-frame pore-scale facial trajectories from PoreTrack3D, including frontal 2D projections and corresponding 3D trajectories displayed with a colored facial point cloud as a spatial reference. To facilitate visual inspection, trajectories of different types are color-coded: green denotes the primary trajectories that are successfully tracked across time, blue indicates the completion-derived trajectories obtained through temporal completion, and red represents supporting trajectories derived from neighboring keypoints. For clarity, only the trajectories over the previous five frames (out of the total 150) are displayed for each frame, and the sequences are arranged from easy to difficult according to the order of dataset construction. The same ordering is consistently applied in all subsequent visualizations and tables.

Since the point clouds reconstructed by COLMAP lack an absolute scale, the world coordinate system was calibrated by computing the average 3D intercanthal distance from dense point clouds of multiple male subjects in the NeRSemble dataset and aligning it with the canonical human intercanthal distance of approximately 33 mm\cite{evereklioglu2001normative}.

\subsection{Metrics}

To accompany the PoreTrack3D dataset, we employ three evaluation metrics to quantitatively assess 3D motion accuracy and temporal robustness: Mean Median Trajectory Error (MTE), Positional Accuracy within Thresholds ($\delta$), and Trajectory Survival Rate (Sur).  
Each metric is computed from the ground-truth trajectories $\{P_{i_t}\}$ and predicted trajectories $\{\hat P_{i_t}\}$, where $i \in [1, n]$ indexes keypoints and $t \in [1, 150]$ indexes frames.

\textbf{Median Trajectory Error (MTE).} This metric measures the typical spatial deviation between predicted and ground-truth trajectories.  
For each trajectory $i$, we compute the per-frame Euclidean distance
\begin{equation}
d_{i_t} = \| P_{i_t} - \hat{P}_{i_t} \|_2 ,
\end{equation}
followed by the median error across time steps,
\begin{equation}
m_i = \mathrm{median} ( d_{i_t} ).
\end{equation}
The overall MTE is obtained by averaging these median errors across all tracked points:
\begin{equation}
\mathrm{MTE} = \frac{1}{n} \sum_{i=1}^{n} m_i .
\end{equation}
This formulation captures typical per-trajectory deviations while suppressing the influence of outlier frames.

\textbf{Positional Accuracy within Thresholds ($\delta$).} To evaluate how consistently each tracker maintains geometric precision, we compute the proportion of points whose prediction error falls below several spatial thresholds $\tau \in \{1, 1.5, 2, 2.5\}$mm .  

For each threshold $\tau$, we compute
\begin{equation}
\delta(\tau) = \frac{\text{Number of points satisfying } d_{i_t} < \tau}{\text{Total number of points } (n \times 150)} .
\end{equation}
This value represents the percentage of all predicted 3D points whose errors are smaller than the chosen threshold.  
The final positional accuracy is obtained by averaging over all thresholds:
\begin{equation}
\overline{\delta} = \frac{1}{|\mathcal{T}|}\sum_{\tau \in \mathcal{T}} \delta(\tau).
\end{equation}

\textbf{Trajectory Survival Rate (Sur).} It measures the temporal stability of trajectory tracking by evaluating how long a trajectory can remain accurate before its first failure \cite{zheng2023pointodyssey, luiten2024dynamic}. For each trajectory, survival is counted from its initialization (frame~1) until the first frame where the tracking error exceeds 3~mm. Formally, for trajectory $i$, we record its valid duration as
\begin{equation}
\ell_i = \min \{\, t \mid d_{i_t} > 3 mm \,\},
\end{equation}
and define Sur as the average normalized lifespan across all trajectories:
\begin{equation}
\mathrm{Sur} = \frac{1}{n} \sum_{i=1}^{n} \frac{\ell_i}{150}.
\end{equation}
This metric reflects the expected proportion of frames a trajectory can be accurately maintained before significant drift occurs.

To complement trajectory evaluation, we also assess multi-view reconstruction quality using standard image-level metrics, including PSNR, SSIM, and LPIPS. These are reported solely for completeness and to provide a consistent comparison with prior 3D dynamic reconstruction benchmarks.

\section{Evaluation on PoreTrack3D}

This section evaluates the performance of state-of-the-art dynamic 3D Gaussian splatting methods on the PoreTrack3D benchmark, outlining the experimental setup and reporting both quantitative and qualitative results. The aim is to assess how existing reconstruction paradigms handle subtle facial dynamics and to establish a unified baseline for future work on millimeter-level 3D tracking.

\subsection{Benchmark Setup}
Prior studies \cite{luiten2024dynamic, duisterhof2023deformgs, zhang2024dynamic} have shown that dynamic 3D Gaussian splatting methods surpass correspondence-based tracking approaches in temporal coherence and trajectory accuracy, demonstrating their effectiveness for 3D point tracking. To provide a comprehensive benchmark and establish reliable baselines for future research, we evaluate four state-of-the-art dynamic scene reconstruction methods based on 3D Gaussian splatting, each representing a different modeling strategy within dynamic 3D Gaussians. All methods are retrained using the dense 3D point clouds generated during dataset construction as initialization, ensuring consistent geometric priors across evaluations. All experiments are conducted on four NVIDIA A6000 Ada GPUs.

We first evaluate two general-purpose dynamic scene modeling methods. Dyn3DGS \cite{luiten2024dynamic} incrementally optimizes the positions and orientations of 3D Gaussians over time under physical constraints, yielding temporally consistent trajectories for each Gaussian. 4DGS \cite{wu20244d} learns a spatiotemporal deformation field via a HexPlane encoder and multi-head decoder, mapping canonical Gaussians to dynamic states at arbitrary timestamps.

Two specialized models are further included to examine performance in more complex, structured settings. GSAvatars \cite{qian2024gaussianavatars} adopts a model-driven design by binding 3D Gaussians to a parametric facial model (FLAME), achieving precise control and high-fidelity reconstruction. Dyn3DGS-robot \cite{zhang2024dynamic} extends Dyn3DGS with a structured control-particle representation for improved temporal consistency and controllability in dynamic tracking. In this evaluation, only its tracking module is used, excluding the learned dynamics prediction component.

All evaluations are performed on eight subjects covering diverse skin tones, genders, ages, and expressions. Each subject contains multiple manually verified trajectories spanning 150 frames, including both 3D landmark and pore-scale keypoints.

\subsection{Performance of Baselines}
\begin{table*}[ht]
    \centering
    \caption{Evaluation of all methods on trajectory tracking and self-reenactment metrics.}
    \label{tab:table2}
    \begin{tabular}{@{}cc ccc@{\qquad}ccc@{\qquad}ccc@{}}
        \toprule
        \multicolumn{2}{c}{} & \multicolumn{3}{c}{\textbf{Self-Reenactment}} & \multicolumn{3}{c}{\textbf{3D Tracking(landmark)}} & \multicolumn{3}{c}{\textbf{3D Tracking(pore-scale)}} \\
        \cmidrule(lr){3-5} \cmidrule(lr){6-8} \cmidrule(lr){9-11}
        \textbf{ID} & \textbf{Method} & \textbf{SSIM} & \textbf{PSNR} & \textbf{LPIPS} & \textbf{MTE} & \textbf{$\boldsymbol{\overline{\delta}}$} & \textbf{Sur} & \textbf{MTE} & \textbf{$\boldsymbol{\overline{\delta}}$} & \textbf{Sur} \\
        \cmidrule{1-11}
        \multirow{4}{*}{\textbf{063}} 
        & Dyn3DGS         & 0.87    & 28.49 & 0.27 & 2.56 & 21.16 & 20.66 & 2.03 & 43.71 & 56.66 \\
        & Dyn3DGS-robot   & 0.77    & 19.25 & 0.39 & 2.81 & 19.08 & 1.33 & 2.81 & 42.40 & 46.67 \\
        & 4DGS            & 0.87    & 31.73 & 0.17 & 1.56 & 61.06 & 81.46 & 2.71 & 37.23 & 38.09 \\
        & GSAvatars        & \textbf{0.98}    & \textbf{39.94} & \textbf{0.01} & \textbf{1.26} & \textbf{71.41} & \textbf{98.00} & \textbf{1.78} & \textbf{65.30} & \textbf{74.33} \\
        \midrule
        \multirow{4}{*}{\textbf{124}} 
        & Dyn3DGS         & 0.87    & 32.21 & 0.28 & 1.31 & 58.33 & 60.33 & 2.01 & 39.07 & 52.95 \\
        & Dyn3DGS-robot   & 0.78    & 18.72 & 0.40 & 2.52 & 30.33 & 6.3 & 2.47 & 43.10 & 48.38 \\
        & 4DGS            & 0.85    & 31.44 & 0.25 & 1.83 & 39.58 & 83.33 & 2.46 & 29.25 & 53.57 \\
        & GSAvatars        & \textbf{0.97}    & \textbf{40.32} & \textbf{0.02} & \textbf{0.81} & \textbf{79.16} & \textbf{96.66} & \textbf{1.64} & \textbf{52.83} & \textbf{79.71} \\
        \midrule
        \multirow{4}{*}{\textbf{056}} 
        & Dyn3DGS         & 0.92    & 33.17 & 0.23 & 1.63 & 54.83 & 27.33 & \textbf{1.52} & \textbf{57.50} & \textbf{69.42} \\
        & Dyn3DGS-robot   & 0.87    & 22.00 & 0.38 & 25.91 & 5.16 & 6.67 & 38.94 & 1.59 & 4.28 \\
        & 4DGS            & 0.91    & 35.50 & 0.19 & 3.02 & 24.50 & 26.00 & 3.82 & 11.64 & 14.28 \\
        & GSAvatars        & \textbf{0.97}    & \textbf{37.33} & \textbf{0.03} & \textbf{1.32} & \textbf{63.16} & \textbf{88.00} & 2.98 & 28.97 & 35.22 \\
        \midrule
        \multirow{4}{*}{\textbf{038}} 
        & Dyn3DGS         & 0.94    & 33.46 & 0.17 & 1.92 & 46.41 & 76.33 & 2.27 & 44.83 & 52.85 \\
        & Dyn3DGS-robot   & 0.89    & 19.67 & 0.28 & 3.65 & 20.25 & 4.66 & 4.08 & 8.64 & 19.23 \\
        & 4DGS            & 0.94    & 35.13 & 0.12 & \textbf{1.21} & \textbf{71.50} & \textbf{81.66} & 4.82 & 14.76 & 33.90 \\
        & GSAvatars        & \textbf{0.98}    & \textbf{39.86} & \textbf{0.01} & 2.44 & 26.83 & 51.33 & \textbf{1.99} & \textbf{45.50} & \textbf{68.16} \\
        \midrule
        \multirow{4}{*}{\textbf{196}} 
        & Dyn3DGS         & 0.90    & 27.80 & 0.26 & 3.72 & 2.33 & 2.66 & 2.16 & 40.67 & 53.52 \\
        & Dyn3DGS-robot   & 0.84    & 18.43 & 0.35 & 2.74 & 9.5 & 24.00 & 3.83 & 27.57 & 33.23 \\
        & 4DGS            & 0.88    & 26.05 & 0.23 & 2.79 & 41.08 & 56.66 & 6.14 & 3.94 & 7.77 \\
        & GSAvatars        & \textbf{0.97}    & \textbf{37.07} & \textbf{0.02} & \textbf{1.41} & \textbf{68.91} & \textbf{100} & \textbf{1.92} & \textbf{47.57} & \textbf{69.43} \\
        \midrule
        \multirow{4}{*}{\textbf{033}} 
        & Dyn3DGS         & 0.93    & 33.26 & 0.24 & \textbf{2.05} & \textbf{39.55} & 29.55 & \textbf{1.61} & \textbf{51.14} & \textbf{66.28} \\
        & Dyn3DGS-robot   & 0.89    & 22.75 & 0.34 & 3.62 & 22.83 & 29.33 & 2.91 & 29.85 & 38.28 \\
        & 4DGS            & 0.94    & 36.68 & 0.13 & 3.43 & 12.50 & 4.00 & 3.22 & 19.42 & 23.23 \\
        & GSAvatars        & \textbf{0.98}    & \textbf{40.20} & \textbf{0.01} & 2.81 & 28.49 & \textbf{50.88} & 2.07 & 43.71 & 54.85 \\
        \midrule
        \multirow{4}{*}{\textbf{264}} 
        & Dyn3DGS         & 0.92    & 31.31 & 0.21 & 1.81 & 48.66 & \textbf{100} & 2.63 & 31.90 & \textbf{50.76} \\
        & Dyn3DGS-robot   & 0.84    & 18.79 & 0.34 & 1.88 & 43.50 & \textbf{100} & 6.31 & 1.07 & 4.28 \\
        & 4DGS            & 0.91    & 33.52 & 0.15 & 1.09 & 69.83 & 84.66 & 3.42 & 18.85 & 24.38 \\
        & GSAvatars        & \textbf{0.97}    & \textbf{39.97} & \textbf{0.01} & \textbf{0.85} & \textbf{89.91} & \textbf{100} & \textbf{2.32} & \textbf{34.38} & 31.61 \\
        \midrule
        \multirow{4}{*}{\textbf{031}} 
        & Dyn3DGS         & 0.96    & 35.09 & 0.16 & 1.72 & 50.91 & 48.33 & 3.78 & \textbf{22.35} & \textbf{22.09} \\
        & Dyn3DGS-robot   & 0.93    & 22.02 & 0.25 & 7.28 & 16.50 & 2.00 & 6.42 & 5.16 & 11.23 \\
        & 4DGS            & 0.96    & 35.54 & 0.08 & 4.73 & 12.41 & 7.00 & 5.63 & 8.00 & 12.28 \\
        & GSAvatars        & \textbf{0.98}    & \textbf{40.75} & \textbf{0.01} & \textbf{1.58} & \textbf{57.27} & \textbf{57.11} & \textbf{3.59} & 16.62 & 16 \\
        \bottomrule
    \end{tabular}%
\end{table*}

Table~\ref{tab:table2} reports the quantitative results of the representative baselines on PoreTrack3D. The primary evaluation focuses on landmark and pore-scale trajectory tracking accuracy, while Self-Reenactment is additionally reported as an auxiliary metric that reflects temporal consistency in reproducing the input motion. Across all trajectory metrics, the results consistently align with the modeling assumptions underlying each method.

Dyn3DGS achieves solid overall performance, serving as a strong baseline for dynamic 3D Gaussian optimization. By directly updating Gaussian parameters under physical regularization, it can capture subtle motion and produces competitive trajectory accuracy. However, the lack of explicit structural priors makes it more susceptible to local drift, especially in regions with complex, high-frequency facial motion.

Dyn3DGS-robot exhibits the lowest scores among all evaluated baselines. The method models dynamics using a reduced set of control points and enforces smoothly varying deformations across time, which leads to more conservative motion updates. While this formulation is effective for maintaining stable global structure, it tends to underfit the high-frequency, subtle displacements present in pore-scale facial motion. As a result, the reconstructed trajectories become overly smoothed, causing reductions in both reconstruction fidelity and point-level tracking accuracy.

4DGS, which models dynamics through a learned continuous deformation field parameterized by a HexPlane encoder, attains reconstruction metrics comparable to Dyn3DGS. Its deformation field produces smooth temporal evolution, which benefits large-scale, low-frequency motions such as landmark trajectories. However, this smoothness makes it less capable of representing sharp, pore-scale deformations.

GSAvatars achieves the best overall performance across reconstruction and tracking metrics. By binding 3D Gaussians to the FLAME mesh, the method ensures consistent alignment between geometry and appearance, enabling stable temporal correspondence. This parametric prior helps preserve structural fidelity and allows the model to capture both global shape changes and detailed surface variations, leading to highly coherent keypoint trajectories.

Across all methods, landmark tracking is consistently more accurate than pore-scale tracking, highlighting the inherent difficulty of modeling irregular, millimeter-scale facial deformations.

\subsection{Qualitative Evaluation}
\begin{figure}[t!] 
    \centering
    \includegraphics[width=1\linewidth]{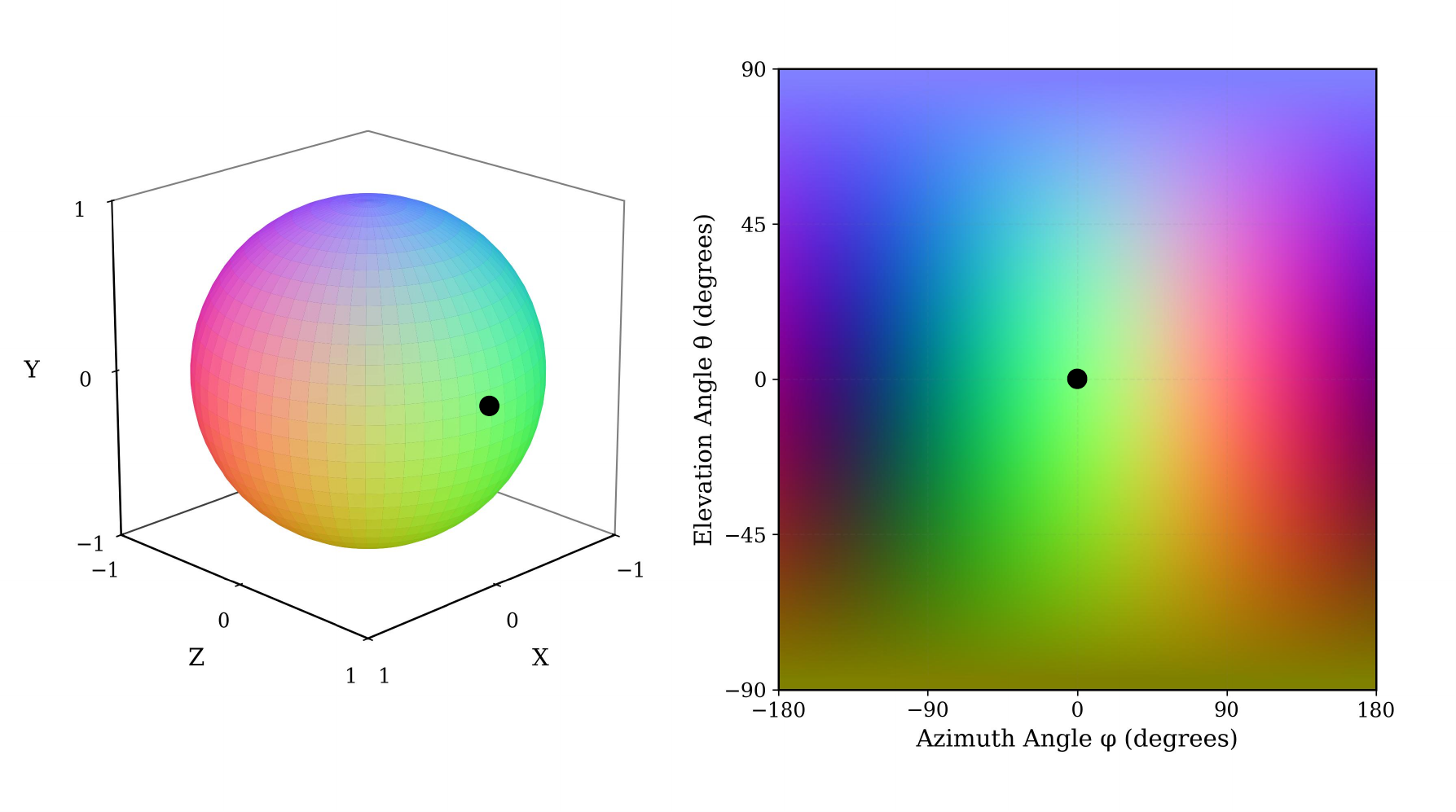}
    \caption{Color Encoding of Motion Direction. Left: 3D unit sphere with surface colors derived from direction vectors via RGB mapping. Right: 2D projection of the sphere, where the horizontal and vertical axes correspond to azimuth and elevation angles, respectively, consistent with the left color encoding. The black dot marks the face’s forward direction, corresponding to the center of the projection.} 
    \label{Fig.7}
\end{figure}
\begin{figure}[t!] 
    \centering
    \vspace{-10pt}
    \includegraphics[width=1\linewidth]{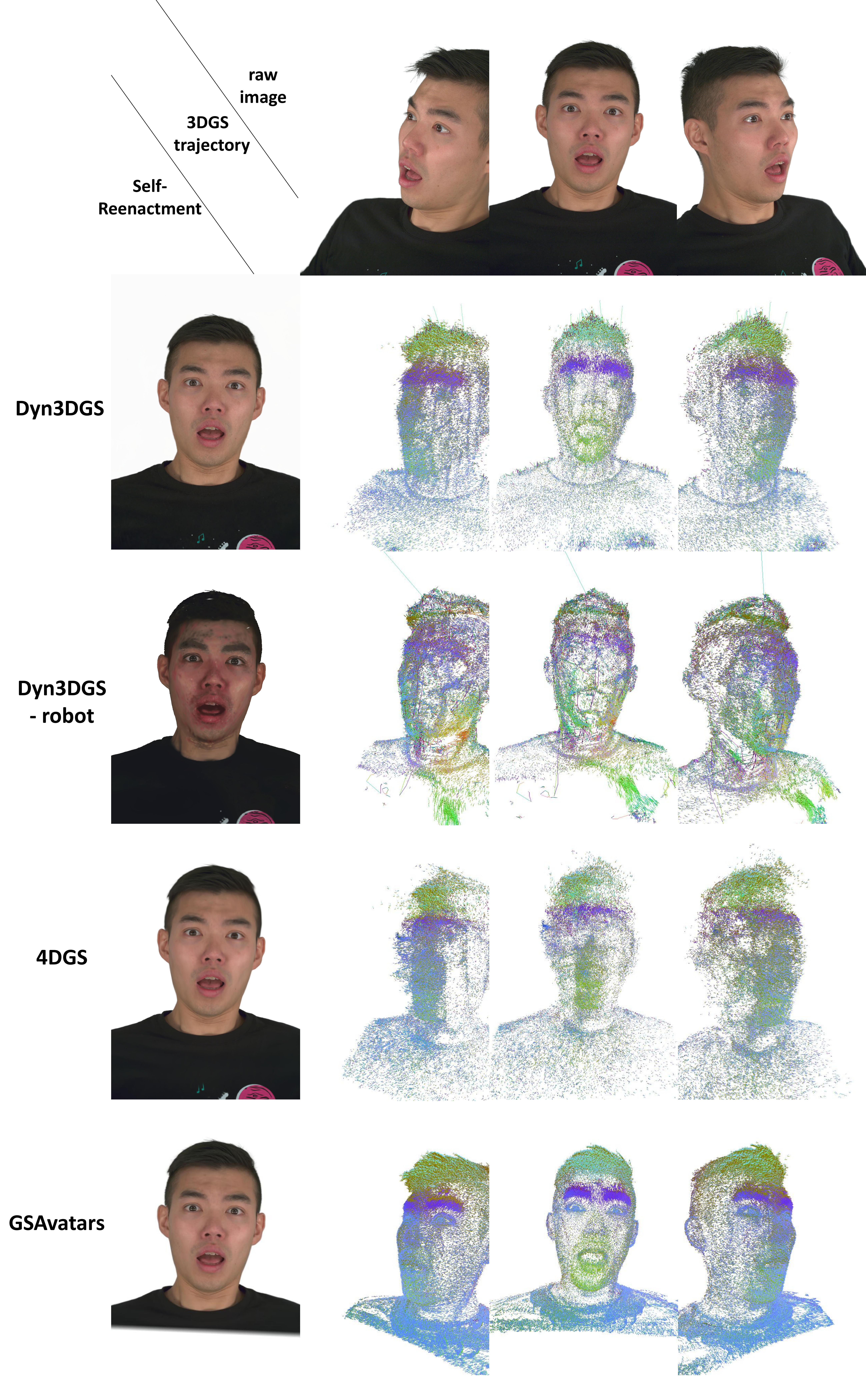}
    \vspace{-10pt}
    \caption{Qualitative comparison. Frontal-view self-reenactment renderings and 3D Gaussian trajectory visualizations from left-profile, right-profile, and frontal viewpoints, showing differences in reconstruction fidelity and motion coherence across methods.} 
    \label{Fig.8}
\end{figure}

In contrast to the previous sections that focus on local patch visualization, this subsection presents a global view of all 3D Gaussian motion trajectories to better illustrate their overall spatial dynamics. To distinguish motion directions across dense trajectories, we design a direction-to-color encoding scheme, as shown in Fig. \ref{Fig.7}. Each unit direction vector \((x, y, z)\) is linearly encoded into the RGB color channels, where $X \mapsto R$, $Y \mapsto B$, and $Z \mapsto G$, with the coordinate range \([-1, 1]\) rescaled to \([0, 1]\). For intuitive interpretation, we define the azimuth angle as positive when the face turns right, and the elevation angle as positive when the face moves upward, aligning with the natural perception of facial motion. This encoding ensures visually continuous and physically interpretable trajectory colors.

Based on this encoding, Fig. \ref{Fig.8} presents qualitative results of four comparative methods on the self-reenactment task, together with their corresponding 3D Gaussian trajectory visualizations over the preceding 10 frames. Consistent with quantitative results, Dyn3DGS-robot produces less coherent motion and lower rendering quality, whereas GSAvatars, benefiting from accurate geometric priors and robust initialization, yields smoother and more stable trajectories, as well as higher rendering quality in the self-reenactment task.

\section{Conclusion}
We have introduced PoreTrack3D, the first benchmark for dynamic 3D Gaussian splatting in pore-scale, non-rigid facial trajectory tracking. The dataset contains over 440,000 trajectories, including more than 52,000 tracks that exceed 10 frames and 68 manually verified tracks spanning the entire 150-frame sequence. Such long-range, densely sampled correspondences are rarely available in existing geometric datasets and provide strong temporal constraints for evaluating subtle, non-rigid surface motion.

By capturing both macroscopic facial landmarks and microscopic pore-scale features, PoreTrack3D bridges the gap between coarse landmark tracking and high-precision surface dynamics. Its data creation pipeline—combining automated extraction with targeted human validation—yields a benchmark that is both scalable and reliable for studying subtle, physically consistent deformation.

Using this benchmark, we establish the first performance baseline for dynamic 3D Gaussian splatting methods, systematically analyzing their ability to model subtle, non-rigid motion. Current 3D Geometry Foundation Models (GFMs) have rarely been evaluated on facial, non-rigid, long-sequence scenarios due to the lack of suitable ground-truth correspondences. PoreTrack3D fills this gap by providing the type of long, dense tracks required for such assessments, thereby enabling—rather than requiring—future work to explore GFMs under this challenging regime.

Looking forward, PoreTrack3D advances the quantitative study of pore-scale facial deformation and provides a valuable supervisory signal for future methods across neural deformation fields, Gaussian-based dynamic modeling, 3D scene-flow estimation, and GFMs. We believe this benchmark establishes an important foundation for precise, interpretable, and physically grounded modeling of 3D facial motion.

\bibliographystyle{IEEEtran}
\bibliography{refs}

\vfill

\end{document}